\documentclass[sigconf]{acmart}

\AtBeginDocument{%
  \providecommand\BibTeX{{%
    \normalfont B\kern-0.5em{\scshape i\kern-0.25em b}\kern-0.8em\TeX}}}

\setcopyright{acmlicensed}
\copyrightyear{2018}
\acmYear{2018}
\acmDOI{XXXXXXX.XXXXXXX}

\acmConference[Conference acronym 'XX]{Make sure to enter the correct
  conference title from your rights confirmation emai}{June 03--05,
  2018}{Woodstock, NY}
\acmISBN{978-1-4503-XXXX-X/18/06}

\usepackage{hyperref}
\usepackage{url}
\usepackage{booktabs}
\usepackage{multirow}
\usepackage{xspace}
\usepackage{graphicx}
\usepackage{float}
\usepackage{subfigure}
\usepackage{makecell}
\usepackage{enumitem}
\usepackage{adjustbox}

\newcommand{\eg}{\emph{e.g.,}\xspace}
\newcommand{\ie}{\emph{i.e.,}\xspace}

\usepackage{color}

\title{Overcoming Pitfalls in Graph Contrastive Learning Evaluation: Toward Comprehensive Benchmarks}

\begin{document}



\author{Qian Ma}
\email{maq5@rpi.edu}
\affiliation{%
  \institution{Rensselaer Polytechnic Institute}
  \streetaddress{110 8th Street}
  \city{Troy}
  \state{New York}
  \country{USA}
  \postcode{12180-3590}
}

\author{Hongliang Chi}
\email{chih3@rpi.edu}
\affiliation{%
  \institution{Rensselaer Polytechnic Institute}
  \streetaddress{110 8th Street}
  \city{Troy}
  \state{New York}
  \country{USA}
  \postcode{12180-3590}
}

\author{Hengrui Zhang}
\email{hzhan55@uic.edu}
\affiliation{%
  \institution{University of Illinois Chicago}
  \streetaddress{1200 West Harrison St.}
  \city{Chicago}
  \state{Illinois}
  \country{USA}
  \postcode{43017-6221}
}

\author{Kay Liu}
\email{zliu234@uic.edu}
\affiliation{%
  \institution{University of Illinois Chicago}
  \streetaddress{1200 West Harrison St.}
  \city{Chicago}
  \state{Illinois}
  \country{USA}
  \postcode{43017-6221}
}

\author{Zhiwei Zhang}
\email{zbz5349@psu.edu}
\affiliation{%
  \institution{Pennsylvania State University}
  \streetaddress{201 Old Main}
  \city{University Park}
  \state{Pennsylvania}
  \country{USA}
  \postcode{16802}
}

\author{Lu Cheng}
\email{lucheng@uic.edu}
\affiliation{%
  \institution{University of Illinois Chicago}
  \streetaddress{1200 West Harrison St.}
  \city{Chicago}
  \state{Illinois}
  \country{USA}
  \postcode{43017-6221}
}

\author{Suhang Wang}
\email{szw494@psu.edu}
\affiliation{%
  \institution{Pennsylvania State University}
  \streetaddress{201 Old Main}
  \city{University Park}
  \state{Pennsylvania}
  \country{USA}
  \postcode{16802}
}

\author{Philip S. Yu}
\email{psyu@cs.uic.edu}
\affiliation{%
  \institution{University of Illinois Chicago}
  \streetaddress{1200 West Harrison St.}
  \city{Chicago}
  \state{Illinois}
  \country{USA}
  \postcode{43017-6221}
}

\author{Yao Ma}
\email{may13@rpi.edu}
\affiliation{%
  \institution{Rensselaer Polytechnic Institute}
  \streetaddress{110 8th Street}
  \city{Troy}
  \state{NewYork}
  \country{USA}
  \postcode{12180-3590}
}

\renewcommand{\shortauthors}{Q.Ma H.Chi H.Zhang K.Liu Z.Zhang L.Cheng S.Wang PS.Yu and Y.Ma}

\begin{abstract}
The rise of self-supervised learning, which operates without the need for labeled data, has garnered significant interest within the graph learning community. This enthusiasm has led to the development of numerous Graph Contrastive Learning (GCL) techniques, all aiming to create a versatile graph encoder that leverages the wealth of unlabeled data for various downstream tasks. However, the current evaluation standards for GCL approaches are flawed due to the need for extensive hyper-parameter tuning during pre-training and the reliance on a single downstream task for assessment. These flaws can skew the evaluation away from the intended goals, potentially leading to misleading conclusions. In our paper, we thoroughly examine these shortcomings and offer fresh perspectives on how GCL methods are affected by hyper-parameter choices and the choice of downstream tasks for their evaluation. Additionally, we introduce an enhanced evaluation framework designed to more accurately gauge the effectiveness, consistency, and overall capability of GCL methods.

\end{abstract}


\keywords{Do, Not, Us, This, Code, Put, the, Correct, Terms, for,
  Your, Paper}



\maketitle

\section{Introduction}\label{sec:intro}

Graph Neural Networks (GNNs)~\cite{li2015gated,gilmer2017neural,GAT} have emerged as a powerful tool for learning representations from graph-structured data, demonstrating remarkable success across various fields including social network analysis~\cite{GNNSocial,fan2019deep, goldenberg2021social}, molecular biology~\cite{GNNMolecular1,GNNMolecular2,GNNMolecular3}, recommendation systems~\cite{GNNRec1,GNNRec2,GNNRec3}, and traffic prediction~\cite{GNNTraf1,GNNTraf2,GNNTraf3}.  By leveraging the rich relational information inherent in graphs, GNNs can capture complex patterns that traditional neural network architectures struggle to process. However, the efficacy of GNNs is largely contingent upon the availability of task-dependent labels to learn meaningful representations~\cite{hamilton2020graph,GNNpowerful,GraphLabel1,GraphSSL}. Unlike labeling in more common modalities such as images, videos, texts, and audio, annotating graphs presents significant challenges due to the complex and often domain-specific nature of graph data~\cite{GraphSSL,MVGRL20,GraphDomain1}. 
This has led to a growing interest in self-supervised learning methods as a means to circumvent the limitations imposed by the need for extensive labeled datasets. Among the various categories of self-supervised learning, graph contrastive learning (GCL) has emerged as a particularly promising approach~\cite{DGI18,GRACE20,GCA21,CCA21,BGRL21,SUGRL22,SFA23}. 

The primary goal of GCL is to pre-train an encoder capable of generating high-quality graph representations without relying on label information, with the hope that these representations can be effectively utilized across a wide array of downstream tasks~\cite{CLCV1,CLCV2,CLNLP,GraphSSL}. However, the current evaluation protocols for GCL methods exhibit significant shortcomings that fail to align with these fundamental goals as detailed below:
\begin{itemize}[leftmargin=*]
    \item GCL methods typically consist of numerous hyper-parameters due to their unique designs in augmentation methods or contrastive objectives. We provide a detailed review of several representative GCL methods in Section~\ref{sec:preliminary}. In existing evaluation protocols, these hyper-parameters are often tuned for each graph dataset, which plausibly involves the use of a validation set from a downstream task. However, in practice, this process contradicts the premise of pre-training without task-specific labels. This would be particularly problematic if the GCL methods are highly sensitive to hyper-parameter configurations.
    \item In existing evaluation procedures, the evaluation of GCL methods is predominantly focused on a single downstream task, usually node classification, conducted on the same dataset used for encoder pre-training. Such a limited evaluation framework may not accurately reflect the encoder's versatility across diverse tasks, potentially leading to misleading comparisons among different GCL methods. 
\end{itemize}

In our study, we conduct a thorough empirical analysis focusing on two crucial elements: (a) how GCL methods' performance is affected by hyper-parameter adjustments in the pre-training phase, and (b) the extent to which a single downstream task can accurately reflect the overall efficacy of GCL methods. Our investigation, detailed in Section~\ref{sec:issues}, confirms that the current evaluation systems are indeed compromised by these concerns. 
Particularly, we observe that some GCL methods are highly sensitive to the settings of hyper-parameters. Although these methods can perform well when hyper-parameters are optimally tuned, their effectiveness can substantially decrease with less-than-ideal settings, making them less practical for pre-training situations where adapting hyper-parameters for each specific downstream task is impractical. Despite this, such methods may still show strong results within the current evaluation models, which do not fully account for the challenges posed by their sensitivity to hyper-parameters~\cite{SFA23,GRACE20}. Additionally, evaluating these methods based on a single downstream task often does not provide a complete picture of their capabilities, as the performance of GCL methods can vary significantly across different tasks. This inconsistency highlights the limitations of current evaluation approaches that focus on singular tasks, thus failing to offer a comprehensive assessment of GCL methods' overall performance.

Hence, to address the identified issues, we propose a new evaluation protocol aimed at a more comprehensive and accurate assessment of GCL methods. This improved protocol incorporates a comprehensive analysis across diverse hyper-parameter configurations and extends the evaluation to include multi-label dataset evaluation. Through these enhancements, our protocol seeks to mitigate the impact of hyper-parameter sensitivity and broaden the scope of evaluation beyond a single task, thereby offering a more comprehensive understanding of GCL method performance. 
\section{Preliminaries}\label{sec:preliminary} 

Graph Contrastive Learning (GCL) focuses on learning superior node representations by distinguishing between node pairs that share similar semantics and those that do not. An anchor node, which can be any selected node in augmented graphs, is paired with semantically similar nodes or graphs (positive examples) to create positive pairs and with dissimilar nodes or graphs (negative examples) to establish negative pairs. GCL models aim to map graph elements such as nodes or graphs into an embedding space where positive pairs are pulled closely together while negative pairs are pushed far apart. To accomplish this, GCL methods are designed with a variety of components. As well-summarized in a prior work \cite{zhu2021empirical}, GCL methods are normally designed differently in the following key components:
\begin{itemize}[leftmargin=*]
    \item \textbf{Data Augmentation:} Data augmentation in GCL aims to create variations of a graph that preserve its major semantic information, thereby helping models to create positive and negative examples. This involves two main approaches: topology transformations and feature transformations. Topology augmentations adjust the graph structure through methods like edge dropping etc. Feature augmentations alter node features by masking to generate diverse representations. The common hypeter-parameters derived from these components are \textit{drop edge rate} and \textit{drop feature rate}.
    \item \textbf{Contrastive Mode:} In GCL, contrasting modes define how an anchor node's similar and dissimilar samples are selected for comparison. There are three primary modes: local-local contrasts nodes at the same level, global-global contrasts entire graph embeddings, and global-local contrasts nodes with their graph-level representation. The choice of mode depends on the task, with node-focused tasks typically using local-local and global-local modes. 
    \item \textbf{Contrastive Objective:}  Contrastive objectives quantify the similarity between positive pairs and the difference from negative pairs. Common objectives include Information Noise Contrastive Estimation (InfoNCE) and Jensen-Shannon Divergence (JSD), which require explicit negative sampling. Other objectives like Bootstrapping Latent Loss (BL) and Barlow Twins (BT) do not require negatives, focusing instead on enhancing positive pair similarity and feature diversity. The temperature hyper-parameter \textit {$\tau$} in InfoNCE is typically adjustable in GCL methods utilizing it as the contrastive objective. 
\end{itemize}
 Building on the general paradigm of GCL outlined above, we next briefly introduce the benchmarked methods, highlighting their unique design components and the corresponding hyper-parameters used to enhance graph representation learning. {\bf (1) DGI:} As a pioneer work of GCL, DGI \citep{DGI18} maximizes mutual information between global graph embeddings and local node embeddings, leveraging JSD as its contrastive loss. Under its design, it has two hyper-parameters to vary, \textit{hidden dimension} and \textit{layer number} of the backbone GCN used to extract representations. {\bf (2) MVGRL}~\cite{MVGRL20} leverages multiple graph views to capture comprehensive structural patterns, applying different GCN configurations for each view to enrich node representations. MVGRL is with the same set of adjustable hyper-parameters as the DGI, but two backbone GCNs are applied on different augmented views. {\bf (3) GRACE}~\cite{GRACE20} focuses on local node-level embedding contrast between two randomly augmented graph views. Each view is copied from the input graph first, next undergoing modifications where a certain percentage of edges and node features are removed and masked, determined by the specified \textit{drop edge rate} and \textit{drop feature rate}, respectively. Different from JSD loss adopted DGI and MVGRL, the adopted contrastive loss for GRACE is InfoNCE and therefore it also has one additional hyper-parameter \textit {$\tau$} to be determined. Further, {\bf (4) BGRL}~\citep{BGRL21} utilizes Bootstrapping Latent loss for graph learning without negative samples, with similar two augmented graph views adjustable with \textit{drop edge rate} and \textit{drop feature rate}. {\bf (5) Graph Barlow Twins}~\citep{barlow22} aims to make the cross-correlation matrix of embeddings from two views as close to the identity matrix as possible, using different hyper-parameters \textit {drop edge rate} and \textit{drop node rate} for graph augmentations on two views. {\bf (6) CCA-SSG} \citep{CCA21} first generates shared node representations from two augmented graph views with similar hyper-parameters mentioned before, then using Canonical Correlation Analysis to maximize correlation between views and decorrelate feature dimensions within each view. CCA-SSG has an additional hyper-parameter \textit{loss reweighting factor} for the trade-off between its losses. {\bf (7) SUGRL}~\cite{SUGRL22} simplifies the contrastive learning process by omitting graph augmentation and similarity determination steps, focusing instead on increasing inter-class differences and decreasing intra-class differences with a triplet loss and an upper bound loss. Given this specific design, there are four unique hyper-parameters for SUGRL, \textit{loss reweighting factor}, \textit {iteration number} for its unique negative samples generation module, \textit {margin parameter}, a hyper-parameter of the triplet loss that penalizes a small gap in distances between a positive pair and a negative pair, and a \textit{bound parameter}, a non-negative tuning parameter used in the upper bound loss. Recently, {\bf (8) COSTA}~\citep{COSTA22} is proposed to address biases in graph augmentation by employing feature augmentation, allowing for better control over data distribution in the latent space. COSTA has \textit {$\tau$} for its InfoNCE contrastive loss and \textit {drop edge rate} and \textit{drop node rate} for two different augmented views. In addition, {\bf (9) SFA}~\citep{SFA23} proposed spectral feature augmentation for GCL, which is designed to re-balance the feature spectrum by iteratively removing low-rank information from the feature matrix. Therefore, apart from \textit {drop edge rate} and \textit{drop node rate}, there are two more hyper-parameters for SFA: \textit {$k$} as the number of iterations used in its spectral feature augmentation, and again \textit {$\tau$} in its adopted InfoNCE loss.

\section{Datasets and Evaluation Metrics}\label{sec:settings}

\subsection{Datasets}\label{sec:datasets}
In this study, we utilize a diverse collection of datasets to evaluate the performance of GCL models across different domains and tasks:
\begin{itemize}[leftmargin=*]
\item \textbf{Cora (Cora), Citeseer (Cite), and Pubmed (Pub)}: Provided by~\citet{Planetoid}, these citation networks represent articles as nodes with sparse bag-of-words features. Edges are citations linking articles, and node labels correspond to article types.

\item \textbf{Amazon Computers (Am-Com) and Amazon Photo (Am-Ph)}: Offered by~\citet{Coauthor}, these datasets are parts of the Amazon co-purchase graph. Nodes represent goods, with edges denoting frequent co-purchases. Features are encoded from bag-of-words reviews, and labels represent product categories.
    
\item \textbf{Coauthor CS (Co-CS) and Coauthor Physics (Co-Phy)}: Sourced from the Microsoft Academic Graph, these networks feature authors as nodes, with edges capturing co-authorship relations. Node features are extracted from keywords in the authors' publications, while labels represent their primary research areas.

\item \textbf{Protein-Protein Interaction (PPI)}: As cited in~\cite{PPI}, this dataset's networks correspond to human tissues. Nodes have multiple labels from gene ontology sets, with features including positional and motif gene sets, plus immunological signatures.
    
\item \textbf{PCG, HumLoc, and EukLoc}: Curated by~\citet{MLGNC} from biological datasets for multi-label classification problems. PCG focuses on protein phenotype prediction. HumLoc and EukLoc target the prediction of subcellular locations in humans and eukaryotes, respectively.
\end{itemize}

The detailed statistics of datasets can be found in Table~\ref{tab:dataset} in the Appendix Section A.

\subsection {Evaluation Metrics}\label{sec:metrics}
\textbf{Accuracy} is the most commonly used metric for evaluating multi-node classification tasks. It measures the proportion of correctly predicted instances over the total instances. Specifically, accuracy is calculated as the ratio of the sum of true positives and true negatives to the total number of cases, making it a straightforward and intuitive measure of model performance.

\noindent \textbf{F1 score}, which is essential for assessing model performance in multi-label node classification tasks, harmonizes precision and recall into a single metric. Below is a detailed breakdown for Macro and Micro F1 Score:
\begin{itemize}[leftmargin=*]
    \item \textbf{Macro F1 Score}:
    \begin{itemize}
        \item Computes F1 scores separately for each label and then averages them.
        \item Ensures equal importance is given to each label, which is especially useful for datasets with imbalanced label distributions.
    \end{itemize}
    \item \textbf{Micro F1 Score}:
    \begin{itemize}
        \item Aggregates true positives, false positives, and false negatives across all labels to compute a collective F1 score.
        \item Focuses on overall performance, weighting more heavily towards labels that appear more frequently.
    \end{itemize}
\end{itemize}

\section{Issues of Existing Evaluation Protocol} \label{sec:issues}
The primary objective of graph contrastive learning is to pre-train an encoder capable of producing high-quality graph representations without label information, with the anticipation that these representations can be efficiently applied to a diverse range of downstream tasks. However, the existing evaluation protocol for graph contrastive learning methods deviates from the aforementioned goals, thus inadequately assessing their efficacy. Specifically, existing evaluation protocols have the following deficiencies:
\begin{itemize}[leftmargin=*]
    \item As elucidated in Section~\ref{sec:preliminary}, GCL methods typically involve numerous hyper-parameters during the pre-training stage. In the current evaluation protocol, these hyper-parameters are typically optimized for each graph dataset~\cite{GRACE20,SFA23}. The hyper-parameter selection is plausibly carried out with a validation set of a downstream task---an approach that is ideally not assumed during the pre-training phase of the encoder. This is essentially problematic especially if the encoders are sensitive to the hyper-parameters.
    \item The existing evaluation protocol predominantly concentrates on a single downstream task, i.e., node classification on the same dataset the encoder is trained. Such a constrained assessment framework may inadequately capture the encoder's performance across the entire task space, rendering comparisons between different GCL methods potentially misleading. In particular, if the downstream task does not effectively represent the diverse range of possible applications, the basis for comparing GCL methods becomes flawed, as it fails to reflect the true generality and adaptability of the encoder's learned representations. 
\end{itemize}
In this section, we undertake empirical experiments to substantiate the aforementioned concerns. 
Our experiments seek to investigate the following two research questions:  
\begin{itemize}[leftmargin=*]
    \item What is the extent of sensitivity of GCL methods to the hyper-parameters during the pre-training stage? 
    \item Is a single downstream task representative enough to comprehensively assess the performance of GCL methods? 
\end{itemize}

\subsection{Sensitivity to Hyper-Parameters}

In this section, we aim to investigate the sensitivity of the GCL methods to the hyper-parameters in the pre-training stage. Specifically, we intend to commence our investigation by undertaking a hyper-parameter tuning process for each GCL method, utilizing the validation set of the downstream task by the conventional model selection procedure. This approach will not only facilitate gaining a deeper understanding of the capabilities inherent within GCL methods but also establish baselines for assessing their sensitivity to hyper-parameters---we will compare their performance under optimally tuned hyper-parameters against the outcomes derived from suboptimal hyper-parameter configurations. 
\subsubsection{Settings}\label{sec:sensitivity_setting}
In this part, we delineate the general experimental settings for our investigation.

\noindent{\bf Search Space of Hyper-Parameters.} As explicated in Section~\ref{sec:preliminary}, various GCL methods typically involve distinct numbers of hyper-parameters, thereby yielding diverse hyper-parameter search spaces for these GCL methods. Therefore, given the computational expense associated with GCL methods, it is impractical and inequitable to tune all hyper-parameters exhaustively. 
Instead, we advocate to randomly sample $20$ combinations of hyper-parameters from the search space of each method. 
Subsequently, the "optimal" set of hyper-parameters is selected from these sampled combinations for each method. Thus, the selected hyper-parameters in this work differ from the ones reported in existing works~\cite{GRACE20,SFA23}. These $20$ combinations of hyper-parameters can be found along with our code implementation~\footnote{https://anonymous.4open.science/r/KDD24-BGPM-A54E}. 
Specifically, throughout the pre-training phase of each encoder, we save encoder versions at predetermined epochs, namely at $50, 100, 500, 1,000$, and $10,000$ epochs. 
This systematic approach enables the evaluation of the encoder's performance at different stages of training without imposing additional computational burdens, requiring only minimal storage space for the saved models.

\noindent \textbf{Pipeline}
Our evaluation procedure adheres to the pipeline employed in previous works~\cite{DGI18,GRACE20,GCA21,PyGCL}. Specifically, we fix the pre-trained encoder and tune a linear task head for the downstream node classification task \eg a linear multi-class classifier utilizing the node representations from the encoder as input. The performance of the downstream node classification task is considered as the encoder's performance \ie the performance of the GCL method. 

\subsubsection{Results analysis}\label{sec:potential}

We conduct experiments on 7 multi-class node classification datasets, which are introduced in Section~\ref{sec:settings}. We present and analyze the results as follows.

\begin{figure}[h]
    \centering
    \includegraphics[width=0.9\linewidth]{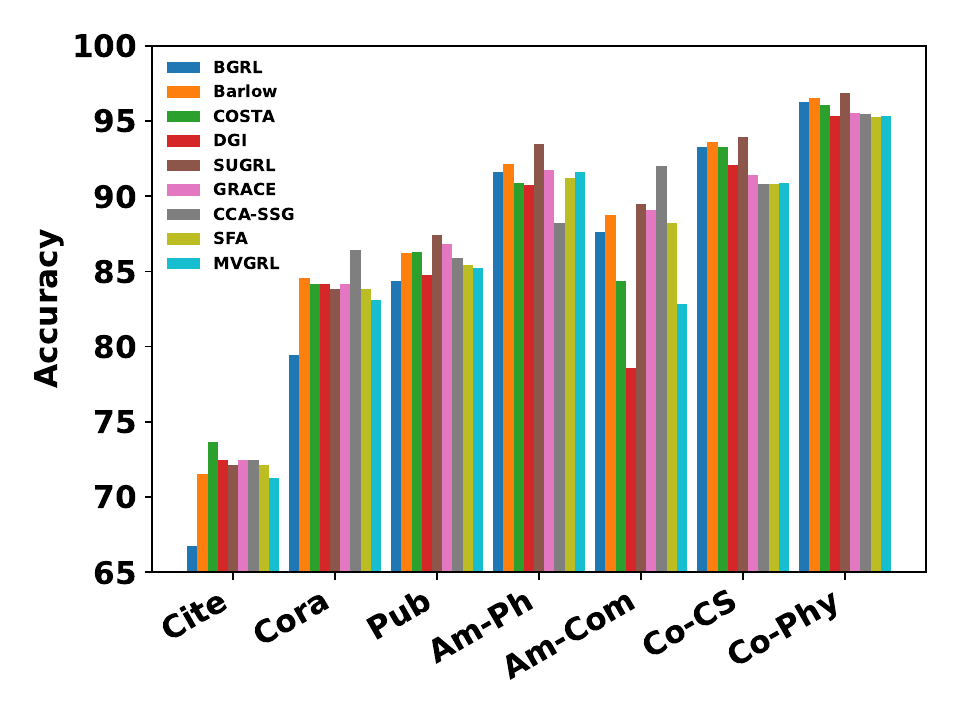}
    \caption{Performance with optimal hyper-parameters. }\label{fig:multi-class-potential}
\end{figure}

\noindent{\bf The Capability of GCL Methods.} We illustrate the capability of GCL methods with their corresponding ``optimal'' hyper-parameters selected from the pre-defined $20$ combinations. The ``optimal'' performance of $9$ representative methods across $7$ datasets are shown in Figure~\ref{fig:multi-class-potential}. 

As shown in Figure~\ref{fig:multi-class-potential}, we observe that all methods are capable of achieving quite strong performance across all datasets with the ``optimal'' hyper-parameters. This observation is consistent with the results reported in the existing works~\cite{SFA23}. Moreover, the performance discrepancies among different methods are often marginal. However, the rankings between them occasionally diverge from existing results. This is mainly due to the different hyper-parameters adopted, which to some extent underscore the sensitivity of the GCL methods to the hyper-parameters. Note that in this experiment, our objective is not to replicate the results from previous studies but rather to demonstrate that all GCL methods can achieve robust performance when appropriately tuned. 

\begin{figure}[h]
    \centering
    \includegraphics[width=0.85\linewidth]{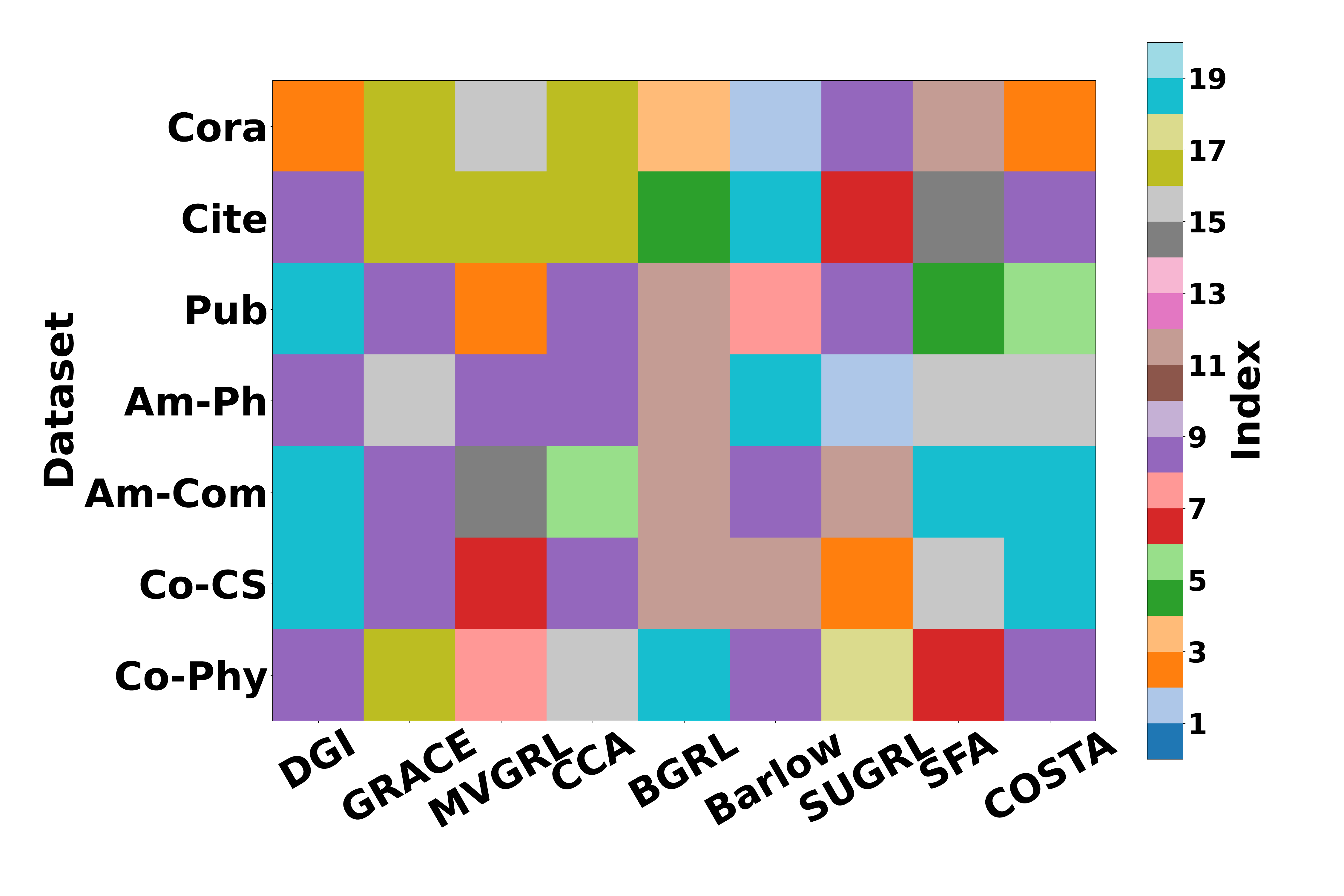}
    \caption{Index of the optimal set for different methods on different datasets. }\label{fig:multi-class-index}
\end{figure}

\noindent{\bf The Selected Hyper-parameters.} In order to investigate how the GCL methods are sensitive to hyper-parameters, we first analyze the selected optimal set of hyper-parameters for different methods across various datasets. In particular, we document the indices of the selected sets of optimal hyper-parameters across all datasets for each method. We illustrate the hyper-parameter selection process for all datasets in Figure~\ref{fig:multi-class-index}, where various indices are represented by different colors in the heatmap.

\begin{figure}[h]
    \centering
    \includegraphics[width=0.85\linewidth]{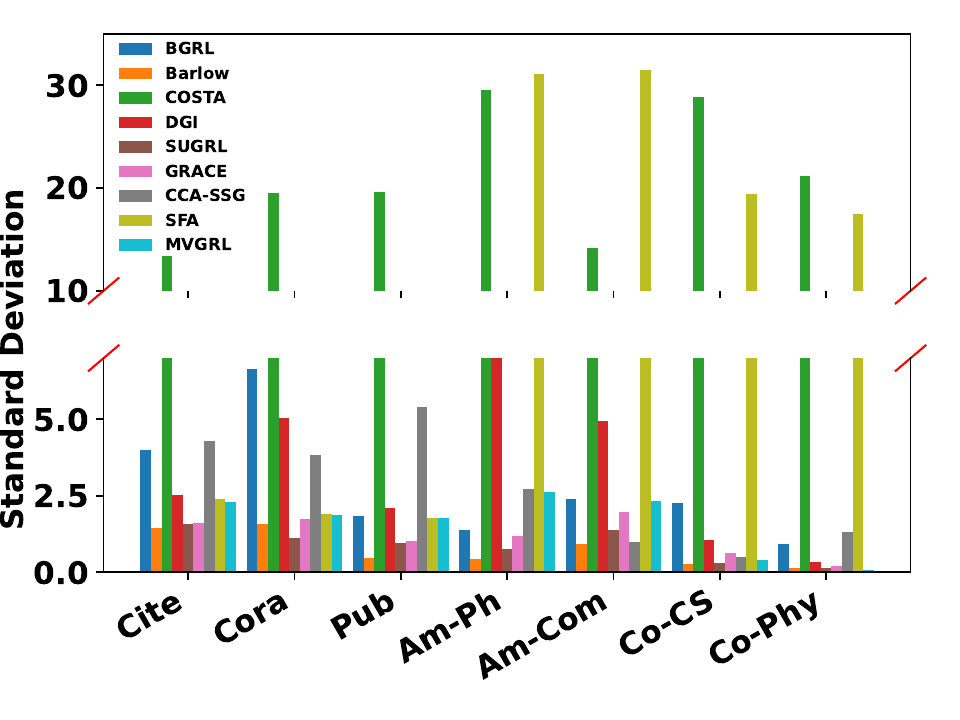}
    \caption{Standard deviation across $20$ sets of hyper-parameters. }\label{fig:multi-class-variance}
\end{figure}
As depicted in Figure~\ref{fig:multi-class-index}, for all methods, there is no dominating set of hyper-parameters valid for all datasets, i.e., different datasets prefer different hyper-parameters. 
These findings underscore the sensitivity of GCL methods to hyper-parameters, indicating the impracticality of employing identical hyper-parameter settings across all datasets, even for the same method. 
Such observations are consistent with existing works~\cite{SFA23,GRACE20}, which have shown that GCL methods often require varied hyper-parameter configurations when applied to different datasets.

\noindent{\bf Variance Across Different Hyper-parameters.}
The preceding results also indicate that the methods are sensitive to hyper-parameters. In this part, we aim to further understand how significantly the GCL methods are affected by the hyper-parameters. In particular, for each method, we assess the downstream performance for the $20$ pre-trained encoders corresponding to the $20$ combinations of hyper-parameters. Subsequently, we compute the standard deviation of these downstream task performances for each method across all datasets. The results are presented in Figure~\ref{fig:multi-class-variance}, where the key observations emerge.
\begin{itemize}[leftmargin=*]
    \item The standard deviation is contingent upon the method, indicating varying degrees of sensitivity among different methods. Notably, COSTA and SFA demonstrate considerably higher instability compared to other methods, while Barlow and SUGRL consistently yield stable results.
    \item The standard deviation is influenced by the dataset utilized for the evaluation, suggesting varying levels of stability across datasets for the same method. For instance, SFA exhibits stability on datasets such as Citesser, Core, and Pubmed but displays pronounced instability on other datasets. Conversely, Barlow exhibits greater stability on Coauthor-CS and Physics datasets, albeit with relatively lower stability on Citeseer and Cora datasets.
\end{itemize}

\noindent{\bf Summary.}
In conclusion, the effectiveness of most GCL methods is notably influenced by the selection of hyper-parameters. When the hyper-parameters are properly optimized, most GCL methods are capable of delivering satisfactory performance for a given downstream task. However, the practical challenge arises from the fact that tuning GCL methods to perfection is often not viable, primarily because downstream tasks are not predefined during the pre-training phase. Therefore, \emph{when evaluating the efficacy of a GCL method, it is crucial to consider its sensitivity to hyper-parameters.} Specifically, GCL methods that exhibit minimal sensitivity to changes in hyper-parameters are generally more desirable. To achieve acceptable performance on downstream tasks, such methods necessitate a reduction in the workload associated with hyper-parameter selection during the pre-training stage.

\subsection{Representativeness of Downstream Tasks}\label{sec:issue2}

In this section, we explore the adequacy of using a single downstream task to evaluate GCL methods.  Our objective is to conduct experiments comparing the performance of various GCL methods across multiple downstream tasks and to assess whether the relative rankings of these methods remain consistent across different tasks.
This investigation seeks to elucidate the representativeness of a single task as a benchmark for assessing the effectiveness of different GCL methods.

\noindent \textbf{Pipeline.} Following  the aforementioned schedule, for the PPI dataset with $121$ labels, we have $121$ binary classification tasks.  Subsequently, we utilize the evaluation setting as described in Section~\ref{sec:sensitivity_setting} to select the best set of hyper-parameters for each task. Thus, after the model selection, for each GCL method, we will have $121$ selected ``optimal'' encoders corresponding to the $121$ tasks. We then proceed to compare the GCL methods on each task with their corresponding ``optimal'' encoders specific to this task. Specifically, for each task, we rank the performance of the $9$ methods. 
Therefore, we compile a total of 121 ranking lists, with each list comprising 9 GCL methods.

\subsubsection{{Results Analysis}}

We present key statistics extracted from the 121 ranking lists in Table~\ref{tab:multi-label-best-lists}. For each GCL method, the ``Min'' and ``Max'' denote its minimum and maximum rankings obtained among the $121$ tasks, respectively. Additionally, the ``Mean" and "Std\_dev" indicate the mean rank and standard deviations of the rank among the $121$ tasks, respectively. 

\begin{table}[h]
\small
\centering
\caption{Optimal Performance on 121 Tasks}\label{tab:multi-label-best-lists}
    \begin{tabular}{@{}ccccc@{}}
    \multicolumn{1}{c}{Model} & Min & Max & Mean  & Std\_dev \\ \hline
    Barlow                    & 1   & 9   & 3.058 & 1.489    \\
    BGRL                      & 1   & 9   & 3.462 & 2.691    \\
    CCA-SSG                   & 2   & 9   & 5.935 & 1.245    \\
    COSTA                     & 1   & 9   & 5.496 & 1.687    \\
    DGI                       & 4   & 9   & 8.545 & 1.113    \\
    GRACE                     & 1   & 8   & 3.025 & 1.474    \\
    MVGRL                     & 1   & 8   & 5.033 & 2.147    \\
    SFA                       & 2   & 9   & 7.074 & 1.337    \\
    SUGRL                     & 1   & 9   & 3.371 & 2.378
    \end{tabular}
\end{table}
The results in table~\ref{tab:multi-label-best-lists} reveal that the rankings of all GCL methods vary across different downstream tasks. 
Each method can exhibit significant performance disparities, achieving high rankings on certain tasks while performing poorly on others, as evidenced by the ``Min'' and ``Max'' rankings.
Moreover, the standard deviation of each method underscores the prevalence of this phenomenon. These findings indicate that no single downstream task is representative enough for the entire downstream task space.
Relying solely on a single downstream task for evaluation may lead to an inaccurate assessment of the effectiveness of GCL methods.

Given the impracticality of hyper-parameter selection for each GCL method, we adopt an alternative experimental approach that involves averaging the performance across $20$ sets of hyper-parameters. 
This methodology aims to mitigate the bias introduced by hyper-parameter sensitivity. Specifically, for each downstream task, we compute the average performance of a GCL method over these $20$ sets of hyper-parameters. 
Subsequently, we apply this average performance metric to rank the GCL methods for each of the 121 tasks, resulting in 121 distinct ranking lists. 
These results are presented in Table~\ref{tab:multi-label-avg}. Similar observations as in Table~\ref{tab:multi-label-best-lists} can be made from Table~\ref{tab:multi-label-avg}, further confirming that a single task is not sufficiently representative.

\begin{table}[h]
    \small
    \centering
    \caption{Average Performance on 121 Tasks}\label{tab:multi-label-avg}
        \begin{tabular}{@{}ccccc@{}}
        \multicolumn{1}{c}{Model} & Min & Max & Mean  & Std\_dev \\ \hline
        Barlow                                      & 1             & 9             & 3.058      & 1.490     \\
        BGRL                                        & 1             & 9             & 3.463      & 2.691     \\
        CCA-SSG                                     & 2             & 9             & 5.934      & 1.245     \\
        COSTA                                       & 1             & 9             & 5.496      & 1.687     \\
        DGI                                         & 3             & 9             & 8.545      & 1.113     \\
        GRACE                                       & 1             & 8             & 3.025      & 1.474     \\
        MVGRL                                       & 1             & 8             & 5.033      & 2.147     \\
        SFA                                         & 2             & 9             & 7.074      & 1.337     \\
        SUGRL                                       & 1             & 9             & 3.372      & 2.378    
        \end{tabular}
\end{table}

\noindent {\bf Summary.} It is evident that relying on a single downstream task for comparing different GCL methods is not sufficient to obtain a comprehensive understanding of their performance. Therefore, \emph{when evaluating the efficacy of GCL methods, it is critical to conduct comprehensive experiments on multiple downstream tasks}.

\section{The Improved Evaluation Protocol}
To address the aforementioned issues, we introduce an improved protocol for a more comprehensive evaluation. This protocol is designed to address the limitations of hyper-parameter sensitivity and the narrow focus on single downstream tasks. The improved evaluation framework consists of the following components.

\begin{itemize}[leftmargin=*]
    \item {\bf Comprehensive Analysis Across Diverse Hyper-parameter Configurations.} 
   We advocate for evaluating GCL methods based on their performance across a variety of hyper-parameter configurations. This approach includes analyzing both the average performance and the variability (or variance) of these performances. By considering the variance, we gain insights into the stability of the GCL methods under different hyper-parameter settings, providing a more holistic view of their effectiveness. 
    \item {\bf Extension to Multi-label Dataset Evaluation.} Our protocol expands the scope of evaluation to include multi-label datasets, which allows us to simulate an environment with multiple downstream tasks. Evaluating GCL methods in this context offers a richer perspective on their capacity to generalize and adapt to diverse downstream tasks.
\end{itemize}

\noindent{\bf General Settings.} Our evaluation encompasses both multi-class and multi-label datasets to ensure a comprehensive analysis of GCL methods. For multi-class datasets, accuracy serves as the primary evaluation metric, offering a straightforward measure of a model's predictive performance. To address the inherent variability and sensitivity of GCL methods to hyper-parameter selection, we undertake an extensive analysis based on the average and variance of performance metrics across $20$ randomly sampled sets of hyper-parameters. For multi-label datasets, our evaluation framework employs both Micro-F1 and Macro-F1 scores as metrics as introduced in Section~\ref{sec:settings}. Adhering to a similar methodology as applied in the multi-class dataset evaluation, we also derive insights from the average and variance of these two metrics across $20$ randomly sampled sets of hyper-parameters for each GCL method.

\subsection{Evaluation with Multi-class Classification\label{sec:multi-class-evaluation0}}

\begin{figure}[h]
    \centering
    \includegraphics[width=0.85\linewidth]{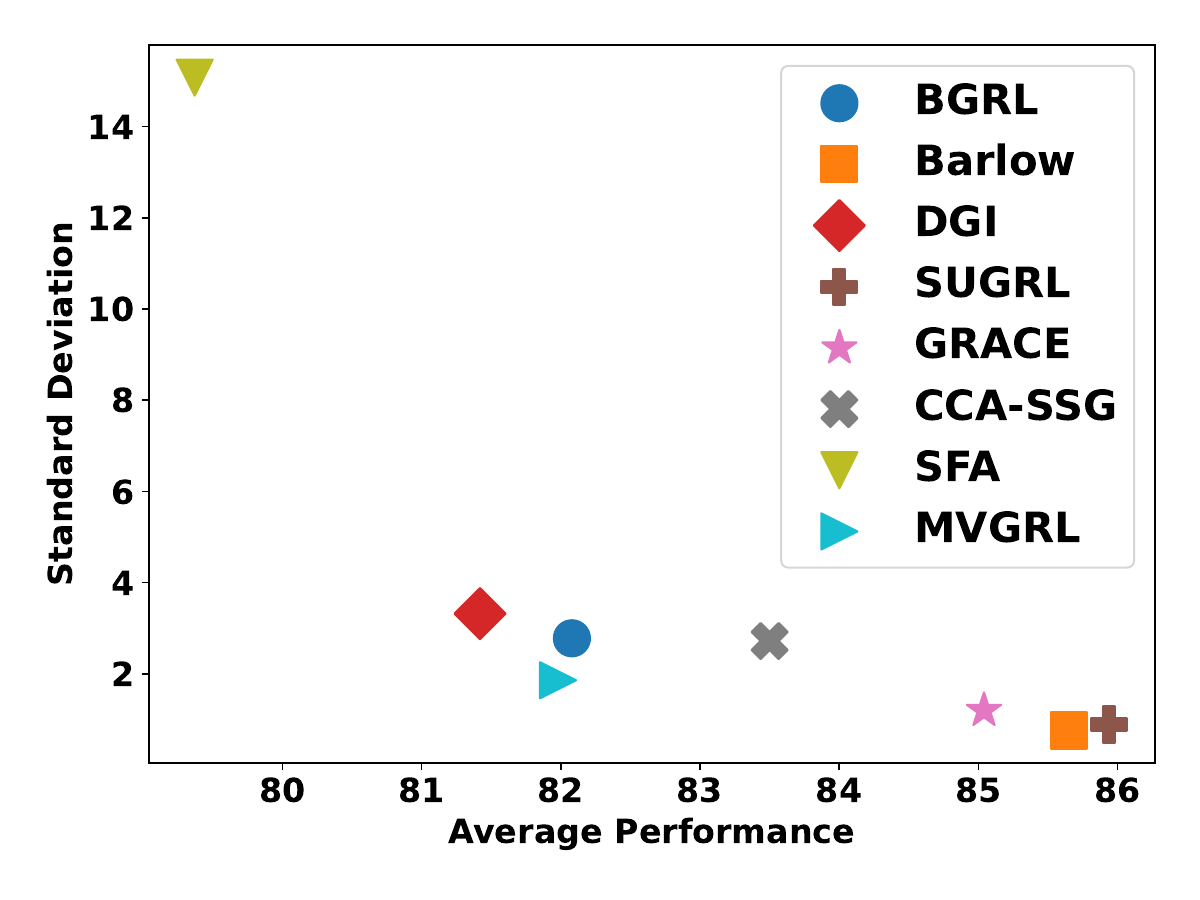}
    \vspace*{-5mm}
    \caption{ Results on multi-class classification datasets.}\label{fig:multi-class-micro}
    \vspace*{-5mm}
\end{figure}

We conduct evaluation procedures for each GCL method on all $7$ multi-class node classification datasets introduced in Section~\ref{sec:datasets}. With our proposed evaluation protocol, we aim to assess the general performance and stability of each method on different datasets. Due to space limit, we aggregate the results of all the datasets into Figure~\ref{fig:multi-class-micro} by averaging them \ie the x and y axis indicates the average values of the method's accuracy and the standard deviation across $7$ datasets, respectively. The performance here specifically denotes accuracy as detailed in Section~\ref{sec:metrics}. Here, a higher mean performance signifies superior effectiveness, whereas a lower mean standard deviation points to enhanced stability.

Notably, in Figure~\ref{fig:multi-class-micro}, COSTA has been excluded due to its notable instability, which is consistent with previously reported findings ~\cite{SFA23,COSTA22}. Barlow and SUGRL stand out by achieving the highest rankings in both average performance and stability metrics, showing their effectiveness and robustness. This finding aligns with the effectiveness reported in their respective papers. Surprisingly, DGI demonstrates a more competitive performance than might be anticipated, considering it as an earlier work often reported to underperform in other Graph Contrastive Learning (GCL) methods. Next, we delved deeper into the discrepancies between our results and those previously published in the literature~\cite{COSTA22,SFA23}. SFA ~\citep{SFA23} reported that SFA consistently surpassed GRACE and Barlow when utilizing a specific set of hyper-parameters, with SUGRL classified as a moderately performing method. In Section~\ref{sec:potential}, SFA is also observed with comparably good results similar to GRACE. Also, COSTA~\cite{COSTA22} was often highlighted as achieving near-top performance, normally just behind SFA. This is also validated by the observation that COSTA performs relatively well in our results shown in Section \ref{sec:potential}. However, when evaluating across our two aforementioned metrics, GRACE and Barlow not only outperform SFA and COSTA, showing superior effectiveness, but SUGRL also emerges as a top performer. Here, SFA's evaluation was based on dataset-specific hyper-parameters according to~\cite{SFA23}, while hyper-parameters for other baselines follow the same ones in PyGCL~\cite{PyGCL}. Those new findings in our results demonstrate the advantage of our proposed evaluation protocol, which is designed to measure the overall performance and stability of GCL methods in a more realistic way. Overall, our new protocol enables a more comprehensive and robust understanding of a GCL method's capabilities.

\begin{figure}[h]
    {\subfigure[{Macro F1}]{\includegraphics[width=0.49\linewidth]{{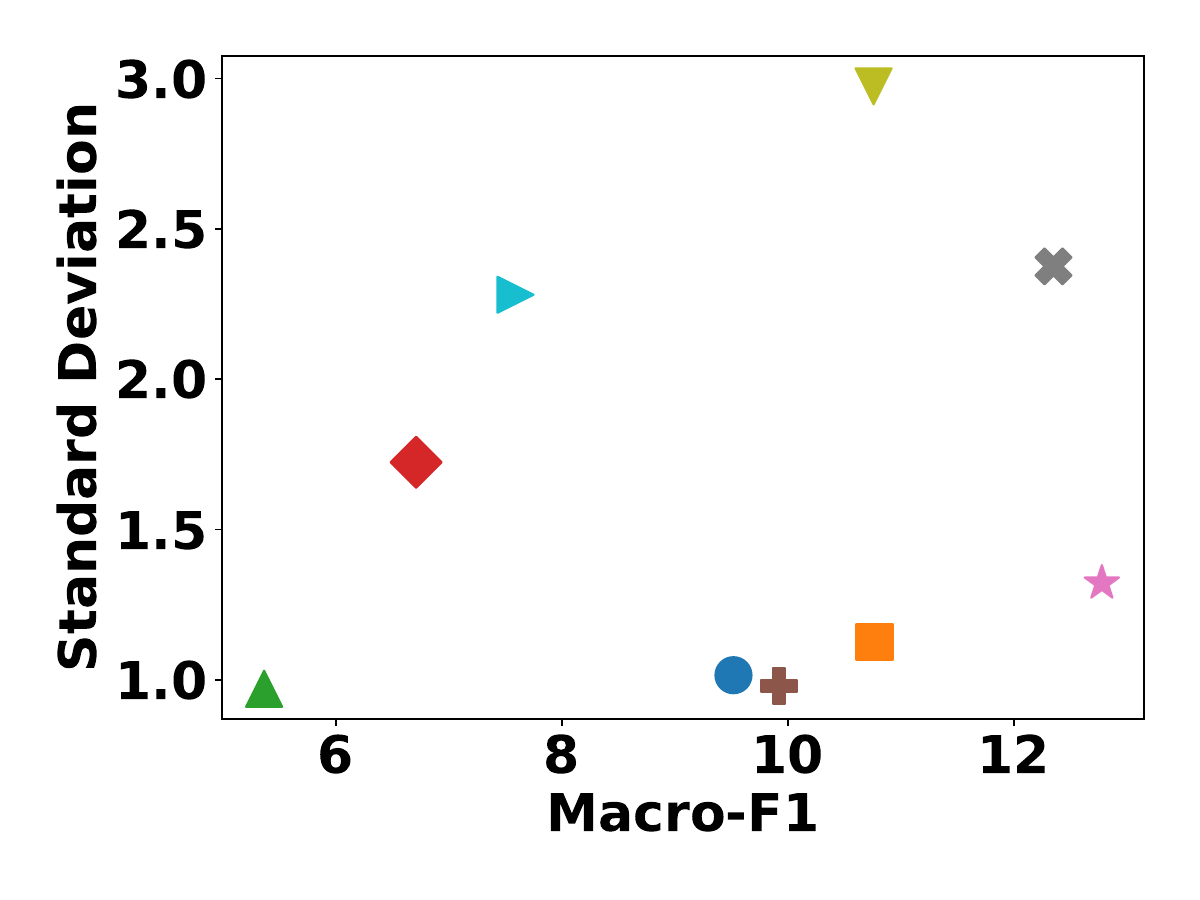}}}}
    {\subfigure[{Micro F1}]{\includegraphics[width=0.49\linewidth]{{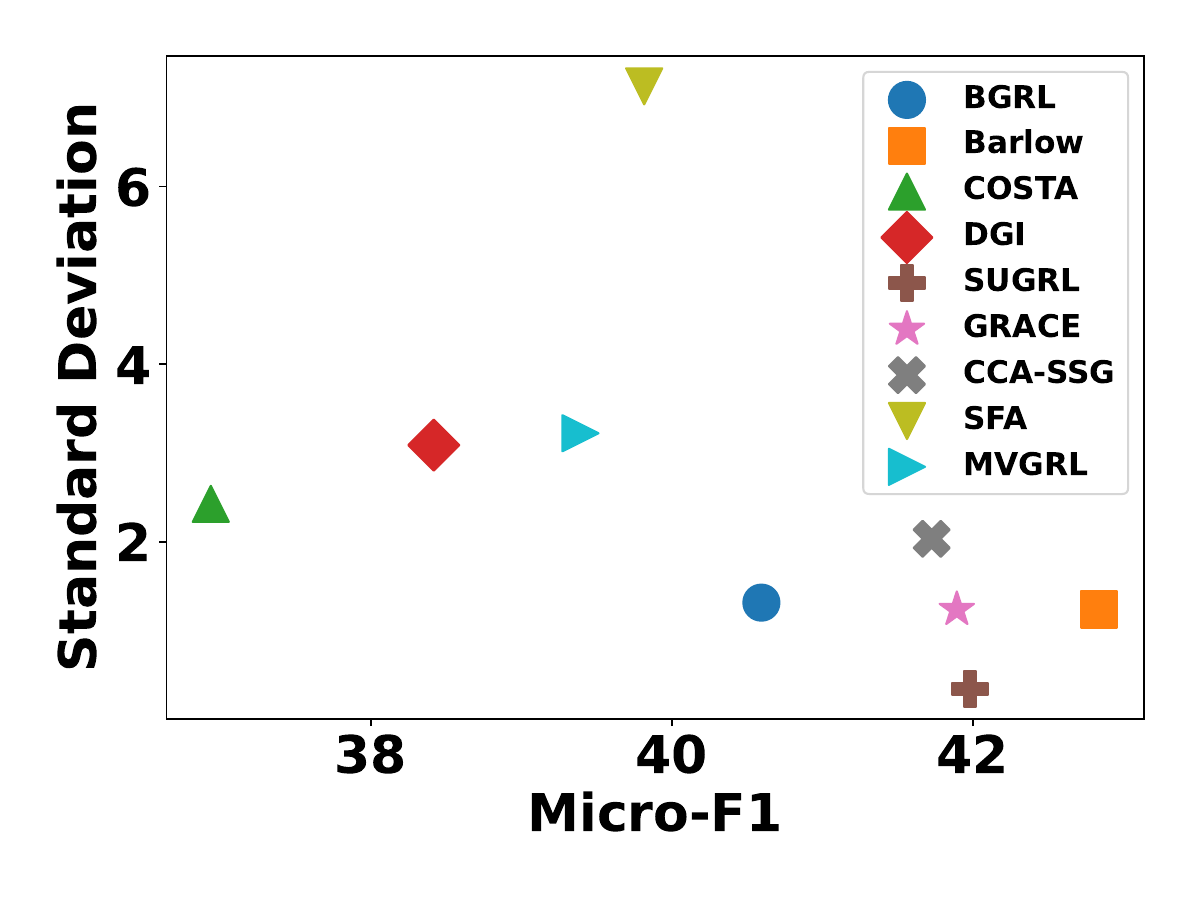}}}}
    \caption{ Results on Eukloc} \label{fig:multi-label-eukloc}
\end{figure}

\begin{figure}[h]
    {\subfigure[{Macro F1}]{\includegraphics[width=.49\linewidth]{{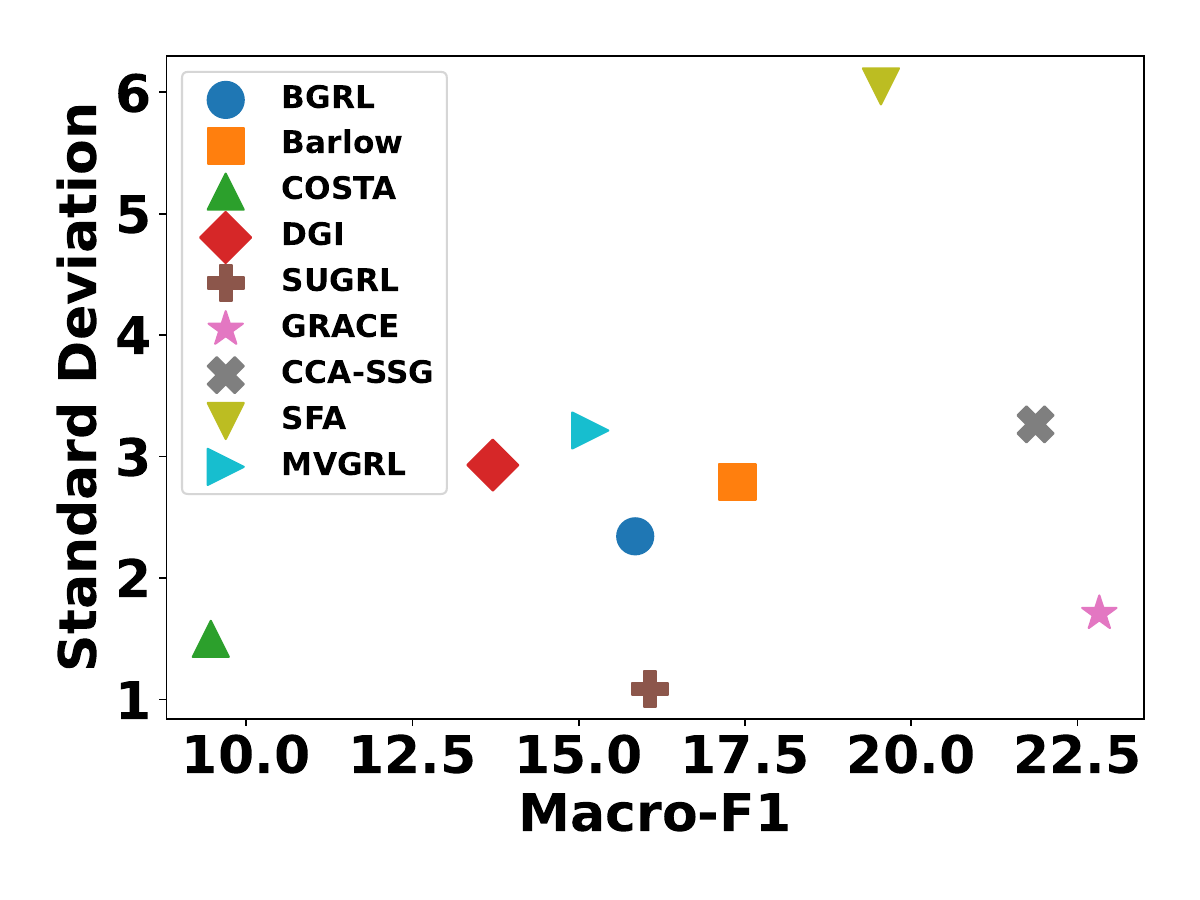}}}}
    {\subfigure[{Micro F1}]{\includegraphics[width=.49\linewidth]{{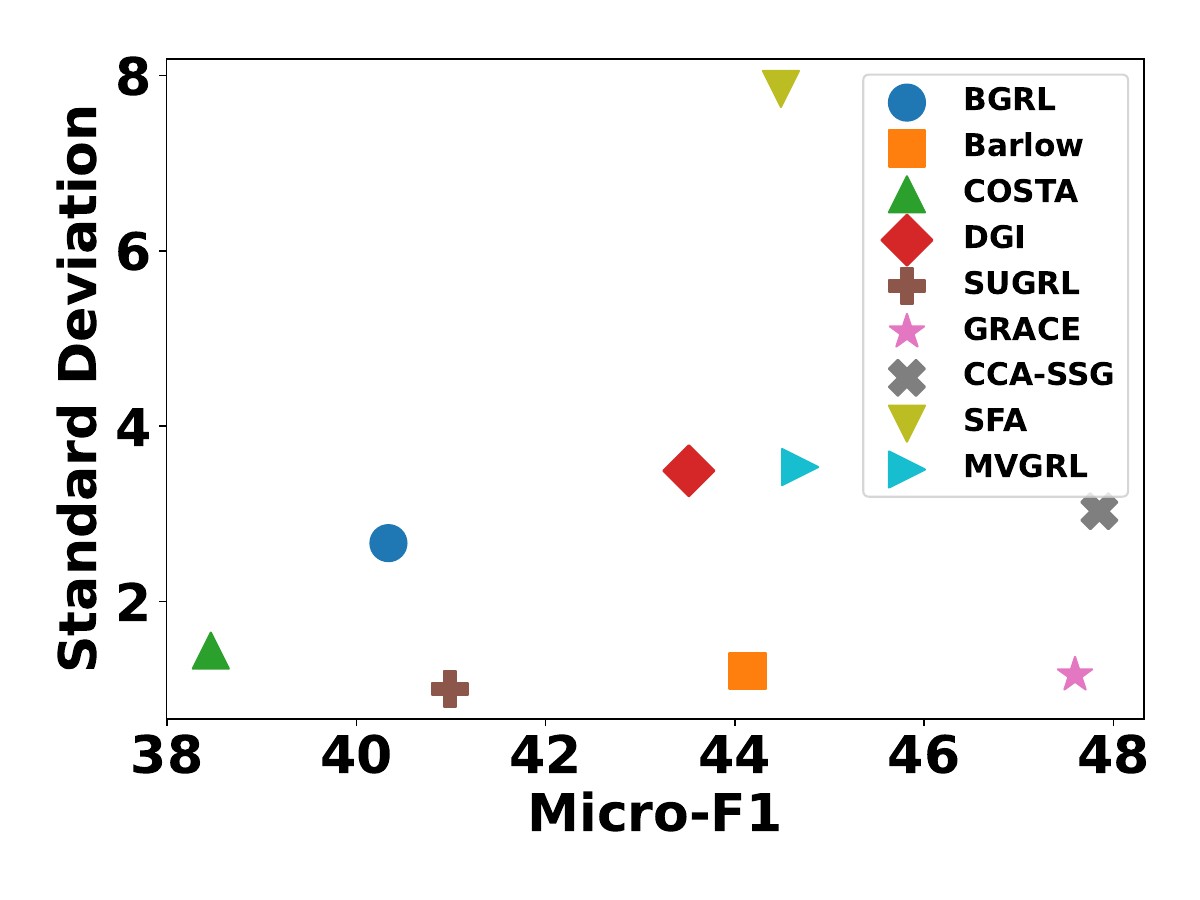}}}}
    \caption{Results on Humloc}\label{fig:multi-label-humloc}
\end{figure}

\begin{figure}[!h]
    {\subfigure[{Macro F1}]{\includegraphics[width=.49\linewidth]{{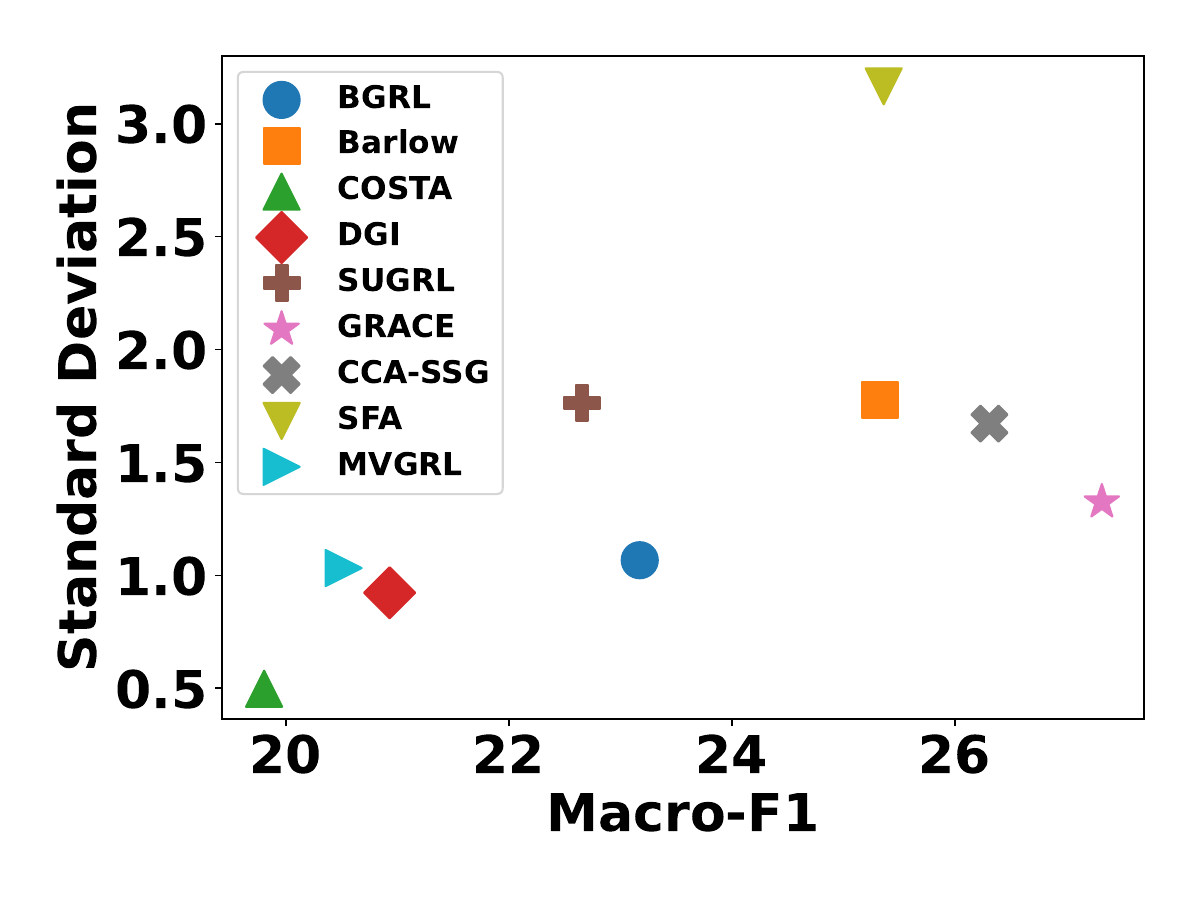}}}}
    {\subfigure[{Micro F1}]{\includegraphics[width=.49\linewidth]{{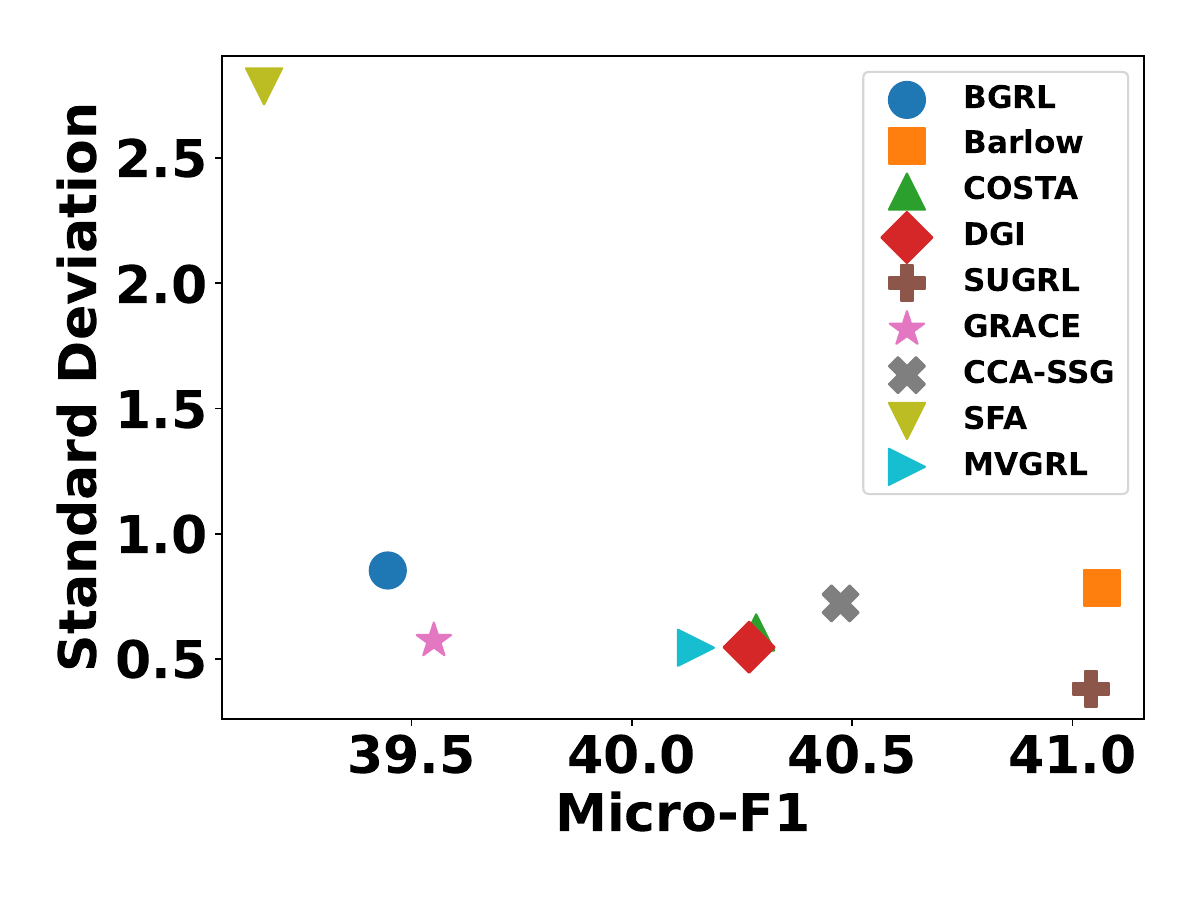}}}}
    \caption{Results on PCG.}\label{fig:multi-label-pcg}
\end{figure}

\subsection{Evaluation with Multi-label Classification}
Following our discussion on the outcomes of multi-class classification, we now shift our focus to the evaluation results obtained from multi-label classification tasks. We conduct the experiments on three datasets Eukloc, Humloc, and PCG, and their corresponding results are presented in Figures~\ref{fig:multi-label-eukloc}, \ref{fig:multi-label-humloc}, and \ref{fig:multi-label-pcg}, respectively. We make some general observations from these figures. Initially, we will discuss findings that align with established benchmarks in multi-class classification, followed by an exploration of divergent results observed in the multi-label classification tasks. 

 In multi-label classification, Grace and Barlow maintain their strong performance and stability, underscoring their robust and consistent capabilities. DGI and BGRL also show a consistent pattern, displaying good stability and moderate performance in both classification contexts. Contrasting the two-fold consistencies mentioned in the prior models, the remaining models exhibit a similar pattern in only one aspect: SUGRL demonstrates notable strength in performance across both evaluation settings, while SFA consistently keeps relatively large standard deviations across all evaluations. Also, CCA-SSG maintains a trend of moderately above-average performance.

 Apart from consistent results, we also observe notable shifts in model performance or overall stability. SFA, previously underperforming in multi-class classification, shows improved results in the current multi-label classification tasks, ranking above average. Conversely, SUGRL, while dominating in multi-class classification, falls to moderate performance levels in the new tasks. In addition, MVGRL sees a slight increase in standard deviation, indicating a change in stability from lower to moderate levels in the multi-label classification. Moreover, CCA-SSG's stability decreases in multi-label classification, with some datasets like the Eukloc showing a notably higher standard deviation in the Macro F1 assessment. 

The observed consistencies across evaluation methods suggest a good level of agreement, indicating that both evaluation methods can, to a certain extent, reflect the overall performance and stability of GCL methods. On the other hand, new observations in multi-label classification tasks highlight the value of incorporating this new multi-label classification evaluation approach, which broadens the evaluation scope and reveals deeper insights into the performance of GCL methods.

\section{Conclusion}

In this paper, we illustrate the intrinsic limitations of the existing evaluation protocol of GCL methods, highlighting its divergence from the overarching goals of self-supervised learning. 
We approach this assessment from two angles: the numerous hyper-parameters during the pre-training phase and the sole evaluation based on a single downstream task.
We verify the prominent issues by investigating the sensitivities of GCL methods to hyper-parameters during the pre-training stage and the representativeness of the downstream task.
Moreover, to rectify the aforementioned issues, we propose an improved evaluation protocol for better benchmarking to comprehensively evaluate the GCL methods.

\newpage
\bibliography{main.bib}
\bibliographystyle{ACM-Reference-Format}
\newpage
\appendix

\section{Supporting materials}
\begin{table}[h]
    \small
    \centering
    \caption{Datasets statistics. }\label{tab:dataset}
    \begin{tabular}{@{}ccccc@{}}
    \textbf{Dataset} & \textbf{\#nodes} & \textbf{\#edges} & \textbf{\#features} & \textbf{\#tasks/\#classes} \\ \midrule
    PCG              & 3k               & 37k              & 32                  & 15                         \\
    HumLoc           & 3.10k            & 18k              & 32                  & 14                         \\
    EukLoc           & 7.70K            & 13K              & 32                  & 22                         \\
    PPI              & 56,944          & 818,716             & 50                  & 121                        \\
    Cora             & 2,708            & 10,556           & 1,433               & 7                          \\
    CiteSeer         & 3,327            & 9,104            & 3,703               & 6                          \\
    PubMed           & 19,717           & 88,648           & 500                 & 3                          \\
    CS               & 18,333           & 163,788          & 6,805               & 15                         \\
    Physics          & 34,493           & 495,924          & 8,415               & 5                          \\
    Computers        & 13,752           & 491,722          & 767                 & 10                         \\
    Photo            & 7,650            & 238,162          & 745                 & 8                         
\end{tabular}
\end{table}

\end{document}